\documentclass[letterpaper, 10 pt, conference]{ieeeconf}  % Comment this line out if you need a4paper

\IEEEoverridecommandlockouts                              % This command is only needed if 
\usepackage[noadjust]{cite}

\usepackage{graphicx} % for includegraphics
\usepackage{comment}
\usepackage{xcolor}
\newcommand\red[1]{\textcolor{red}{#1}}

\usepackage{amsmath}
\usepackage{amssymb} % for mathbb
\usepackage{mathrsfs} % for mathscr
\usepackage{algorithm} % for algorithm
\usepackage{mathtools} % several math tools
\newtheorem{proposition}{Proposition}
\newtheorem{remark}{Remark}

\usepackage{cuted} % for equation over column
\usepackage{siunitx}

\usepackage{ulem}

\usepackage{kotex}

\usepackage{url}

\usepackage{tikz}

\newcommand\copyrighttext{%
  \footnotesize \textcopyright 2023 IEEE.  Personal use of this material is permitted.  Permission from IEEE must be obtained for all other uses, in any current or future media, including reprinting/republishing this material for advertising or promotional purposes, creating new collective works, for resale or redistribution to servers or lists, or reuse of any copyrighted component of this work in other works.}
\newcommand\copyrightnotice{%
\begin{tikzpicture}[remember picture,overlay]
\node[anchor=south,yshift=10pt] at (current page.south) {\fbox{\parbox{\dimexpr\textwidth-\fboxsep-\fboxrule\relax}{\copyrighttext}}};
\end{tikzpicture}%
}
\title{\LARGE \bf
Minimally actuated tiltrotor for perching and normal force exertion
}
\author{Dongjae Lee, Sunwoo Hwang, Changhyeon Kim, Seung Jae Lee, and H. Jin Kim% <-this % stops a space
\thanks{This work was supported in part by Unmanned Vehicles Core Technology Research and Development Program through the National Research Foundation of Korea(NRF) and Unmanned Vehicle Advanced Research Center(UVARC) funded by the Ministry of Science and ICT(NRF-2020M3C1C1A01086411), and in part by Basic Science Research Program through the National Research Foundation of Korea(NRF), funded by the Ministry of Education(NRF-2022R1A6A3A13073267).}%
\thanks{Dongjae Lee, Sunwoo Hwang, Changhyeon Kim, and H. Jin Kim are with the Department of Aerospace Engineering, Seoul National University (SNU), Seoul 08826, South Korea {\tt\small \{ehdwo713, swsw0411, rlackd93, hjinkim\}@snu.ac.kr}}%
\thanks{Seung Jae Lee is with the Department of Mechanical System Design Engineering, Seoul National University of Science and Technology (SEOULTECH), Seoul 01811, South Korea {\tt\small seungjae\_lee@seoultech.ac.kr}}%
}

\begin{document}
\bstctlcite{IEEEexample:BSTcontrol} % to make et.al

\maketitle
\copyrightnotice
\thispagestyle{empty}
\pagestyle{empty}

%%%%%%%%%%%%%%%%%%%%%%%%%%%%%%%%%%%%%%%%%%%%%%%%%%%%%%%%%%%%%%%%%%%%%%%%%%%%%%%%
\begin{abstract}
This study presents a new hardware design and control of a minimally actuated 5 control degrees of freedom (CDoF) quadrotor-based tiltrotor. The proposed tiltrotor possesses several characteristics distinct from those found in existing works, including: 1) minimal number of actuators for 5 CDoF, 2) large margin to generate interaction force during aerial physical interaction (APhI), and 3) no mechanical obstruction in thrust direction rotation.
%Compared to existing works, the proposed tiltrotor has the following properties: 1) minimal actuation, 2) large margin to generate interaction force during aerial physical interaction (APhI), and 3) no mechanical obstruction in thrust direction rotation. 
Thanks to these properties, the proposed tiltrotor is suitable for perching-enabled APhI since it can hover parallel to an arbitrarily oriented surface and can freely adjust its thrust direction. To fully control the 5-CDoF of the designed tiltrotor, we construct an asymptotically stabilizing controller with stability analysis. The proposed tiltrotor design and controller are validated in experiments where the first two experiments of $x,y$ position tracking and pitch tracking show controllability of the added CDoF compared to a conventional quadrotor. Finally, the last experiment of perching and cart pushing demonstrates the proposed tiltrotor's applicability to perching-enabled APhI.

\end{abstract}

%%%%%%%%%%%%%%%%%%%%%%%%%%%%%%%%%%%%%%%%%%%%%%%%%%%%%%%%%%%%%%%%%%%%%%%%%%%%%%%%
\section{Introduction}

%perching-enabled aerial physical interaction이 왜 필요한가
Recently, aerial physical interaction (APhI) has been actively studied in aerial robotics community \cite{ollero2022past} ranging from static structure interaction \cite{bodie2021active} to dynamic structure interaction \cite{lee2021aerial,brunner2022energy,benzi2022adaptive}. Most researches conduct APhI while maintaining hovering flight \cite{lee2021aerial,bodie2021active,byun2021stability,ding2021tilting,brunner2022energy,benzi2022adaptive}, and this requires an aerial robot to persistently generate nonzero thrust, which puts a significant restriction on the maximum allowable interaction force. However, if APhI is available after perching on a contact surface, such endeavor for maintaining hovering flight becomes unnecessary. Accordingly, this perching-enabled APhI has merits over conventional hovering-based APhI in that 1) it is more energy-efficient during APhI and 2) the maximum of allowable interaction force is enlarged. This is because sustaining the hovering maneuver itself takes considerable amount of energy, and a significant portion of the total thrust that is accountable for both gravity compensation and interaction force generation needs to be consigned for the hovering maneuver in hovering-based APhI. Furthermore, if an aerial robot can perch and conduct APhI, such capability reduces the burden on a controller to maintain accuracy and stability during physical interaction, which is still a challenging problem. 

%perching-enabled aerial physical interaction 측면에서 새로운 하드웨어 플랫폼이 필요한 이유
In this study, we consider APhI of exerting substantial normal force to a contact surface whose possible application includes sensor installation, pushing or pulling a movable object, and collaborative aerial transportation. To perform such APhI after perching, thrust vectoring to make the total thrust be perpendicular to a contact surface is first required. Furthermore, for a multirotor to adapt to an uncertain pose of a perching site, orientation and translation should be independently controlled. Such requirements cannot be realized in a conventional multirotor; although it can perch with an additional planning method, for example, \cite{ji2022real}, it cannot rotate its total thrust direction without changing its body orientation. In addition, since a conventional multirotor should rotate its body to induce translation, uncertainty in orientation of a perching site can lead to collision with the perching site for not being able to regulate its terminal velocity. Such inherent limitation motivates a new platform capable of thrust vectoring and independent control of orientation and translation.

\begin{figure}
    \centering
    \includegraphics[width=0.8\linewidth]{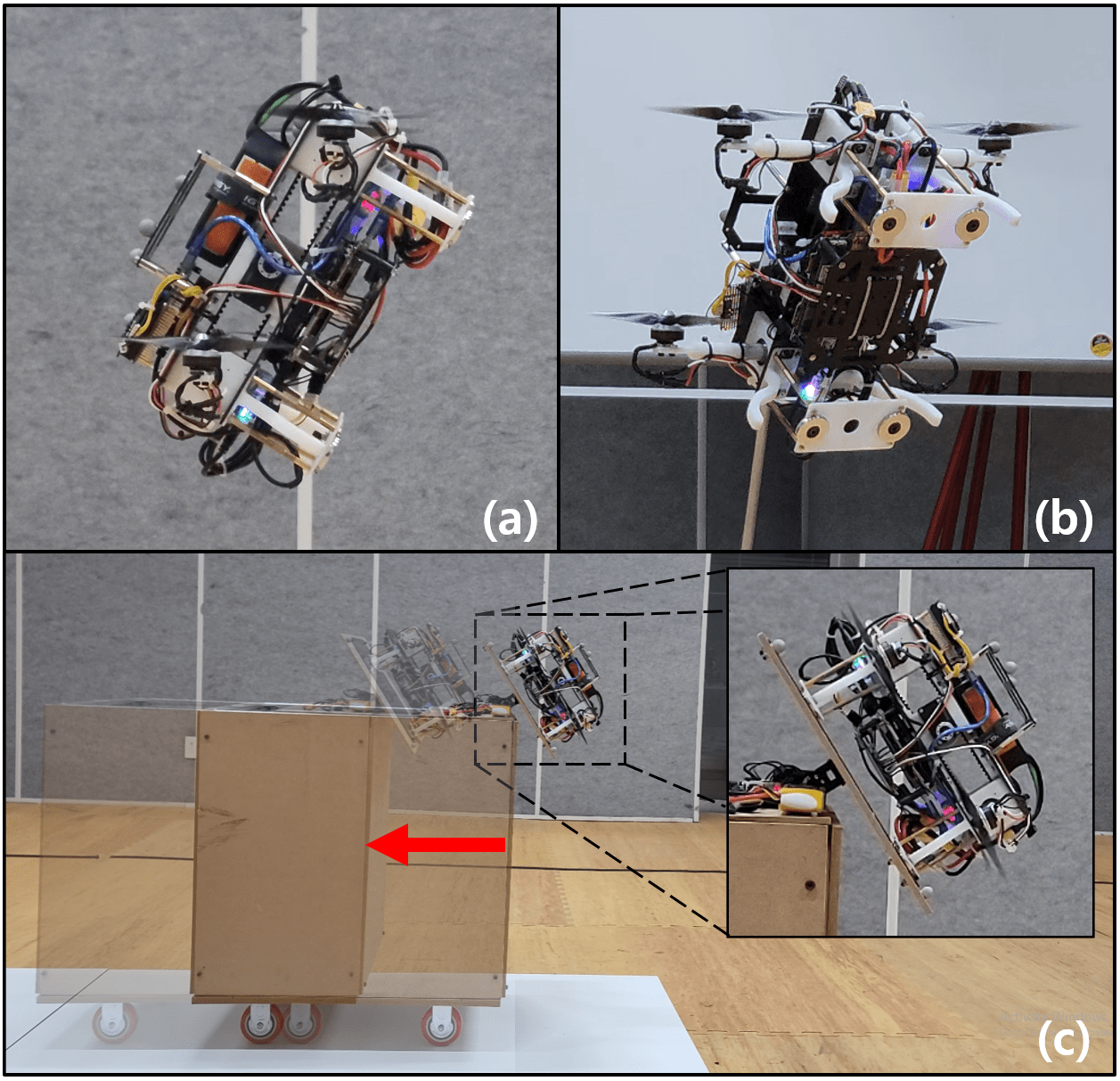}
    \caption{Experimental results of pitch tracking control ((a) and (b)) and cart pushing after perching (c). The perching cite in (c) is installed only to emulate various inclinations.}
    \label{fig:thumbnail}
    \vspace{-0.5cm}
\end{figure}

\begin{figure*}[ht]
    \centering
    \includegraphics[width=0.95\linewidth]{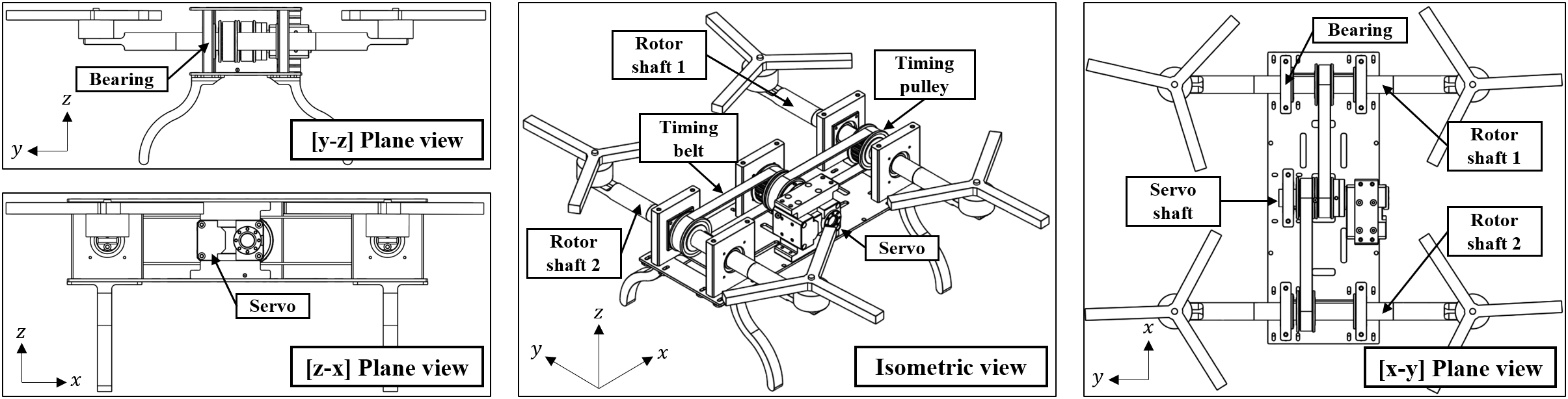}
    \caption{CAD drawings of the tiltrotor. All four rotors rotate by the same angle thanks to the belt-pulley mechanism and a servomotor. Two rotating shafts attached with rotors and one rotating shaft attached to the servomotor are assisted by bearings which are inserted in bearing holders of cuboid shape.}
    \label{fig:cad drawings}
    % \vspace{-0.5cm}
\end{figure*}

The objective of this study is to design and control a minimally actuated five control degrees of freedom (CDoF) quadrotor-based tiltrotor as in Fig. \ref{fig:thumbnail} which can perch on an inclined surface and can exert a normal force to the contact surface. The choice of 5-CDoF comes from the fact that only 2-CDoF in orientation is sufficient for an aerial robot to hover with its orientation parallel to any inclined surface, and thus, the minimum required CDoF for an aerial robot is five. Our objective for minimal actuation results from the need for mass reduction accomplished by efficient mechanism design, and cost efficiency considering future applicability to multi-agent system.

\begin{comment}
\textbf{제안하는 하드웨어의 특징을 아래 단점과 대비하면서 설명하는 것이 더 compact할 듯!}
\begin{itemize}
    \item redundancy: 5-CDoF: \cite{kawasaki2015dual,ding2021tilting,li2023design}, 6-CDoF: \cite{allenspach2020design,brescianini2018omni}
    \item thrust vectoring only in limited range: 6-CDoF: \cite{lee2021fully,zheng2020tiltdrone,ryll2022fast}
    \item reduced efficiency in contact surface-normal force generation: 5-CDoF: \cite{papachristos2014efficient,lee2021caros}
\end{itemize}
\end{comment}

% \subsection{Related work}
Since the minimum required CDoF is five, here we review some multirotor-based platforms with no less than 5-CDoF. Various 5-CDoF multirotors have been proposed and experimentally validated \cite{kawasaki2015dual,ding2021tilting,li2023design,papachristos2014efficient,lee2021caros}. Utilizing either "T" or "H"-shaped multirotor, they all adopt a tilting mechanism to attain additional 1-CDoF in pitch angle. However, they have limitations in that they either are redundant in actuation \cite{kawasaki2015dual,ding2021tilting,li2023design}, or inhere reduced efficiency in surface-normal interaction force generation \cite{papachristos2014efficient,lee2021caros} since $1/3$ of rotors are not connected to the tilting axle.

Multirotors with 6-CDoF have also been widely studied \cite{lee2021fully,zheng2020tiltdrone,ryll2022fast,allenspach2020design,brescianini2018omni,ryll20196d}. Nevertheless, for application to perching-enabled APhI, \cite{lee2021fully,zheng2020tiltdrone,ryll2022fast} suffer from a limited range of thrust vectoring; accordingly, there may exist some cases where desired thrust directions become unachievable during perching-enabled APhI. The full range of thrust vectoring is possible for a platform in \cite{allenspach2020design}, but it is overly redundant in actuation since 12 actuators in total are engaged for 6-CDoF. \cite{brescianini2018omni,ryll20196d} utilize rotors with a fixed tilt angle to achieve full actuation of 6-CDoF; consequently, they might not be applicable to perching-enabled APhI since thrust vectoring is infeasible.

Compared to the other multirotor platforms reviewed above, our proposed platform employs only one additional actuator to obtain 5-CDoF, which makes it minimal in actuation. Furthermore, to achieve maximal unidirectional interaction force margin during APhI, all four rotors can rotate by the same angle. Lastly, no mechanical obstruction in the tilting angle exists which allows a wider achievable thrust direction during APhI. 

Next, to fully control 5-CDoF, we first define a new yaw-compensated coordinate to resolve singularity in controller design which will be further discussed in section \ref{sec:controller design}. Then, to handle underactuatedness due to the platform being 5-CDoF, we decompose the dynamics into two parts and design controllers for both subsystems. Lastly, we propose a motion controller with which asymptotic stability is proved. Compared to existing controllers for 5-CDoF aerial robots \cite{xu2021h,li2023design,lee2021caros,ding2021tilting} where either stability analysis is skipped or roll angle error in translational dynamics is assumed to be negligible, we analyze stability with full consideration of this roll angle error by modeling the closed-loop system as a cascade system. Based on a stability theorem for cascade system, the stability of the entire system is verified.

% \subsection{Contribution}
The contribution of this research can be summarized as follows:
\begin{itemize}
    \item hardware design of a minimally actuated 5-CDoF quadrotor-based tiltrotor with large interaction force margin and no mechanical limit in thrust vectoring angle
    \item controller design and stability analysis to fully control 5-CDoF of the tiltrotor with full consideration of underactuatedness
    \item experimental validation of the proposed tiltrotor and its application to perching-enabled APhI
\end{itemize}

\subsection{Notations}
% \textbf{Notations:}
We use $e_1 \coloneqq [1 \ 0 \ 0]^\top,e_2 \coloneqq [0 \ 1 \ 0]^\top$, and $e_3 \coloneqq [0 \ 0\ 1]^\top$. Furthermore, we define $v_i \in \mathbb{R}$ to be the $i^{th}$ element of a vector $v$. For a column vector $a$ and $b$, $[a;b] \coloneqq [a^\top \ b^\top]^\top$. For a given state variable $x$, we denote its desired value as $x_d$. Lastly, as shorthands of $\cos(\cdot)$, $\sin(\cdot)$, and $\tan(\cdot)$, we use $c(\cdot)$, $s(\cdot)$, and $t(\cdot)$, respectively.

\section{Hardware Design}

While fulfilling the two requirements for \textit{perching-enabled APhI}, which are additional CDoF in orientation and total thrust direction maneuverability, we additionally enforce the following hardware design considerations:
\begin{itemize}
    \item Additional actuators should be used in a minimum number, which is one.
    \item All four rotors should be able to rotate together by the same angle.
    \item The full range of thrust vectoring should be available without any mechanical obstruction.
    %All four rotors should not be mechanically obstructed in thrust direction rotation.
\end{itemize}
Considering both mass reduction and cost efficiency, we restrict the number of additional actuators to one. Next, to secure a large unidirectional interaction force margin during physical interaction, the tiltrotor is designed to be capable of rotating all rotors by the same angle. Compared to cases where not all rotors can be rotated, or some rotors can rotate by different angles only, the interaction force exerted in a single axis can be increased by rotating all rotors by the same angle. Finally, the last design consideration is imposed to achieve both wider achievable thrust direction and large interaction force margin during APhI. If there exists mechanical obstruction to any of the four rotors, interaction force may not be generated in the desired direction or may be reduced in magnitude at certain perching angles.

\begin{figure}
    \centering
    \includegraphics[width=0.75\linewidth]{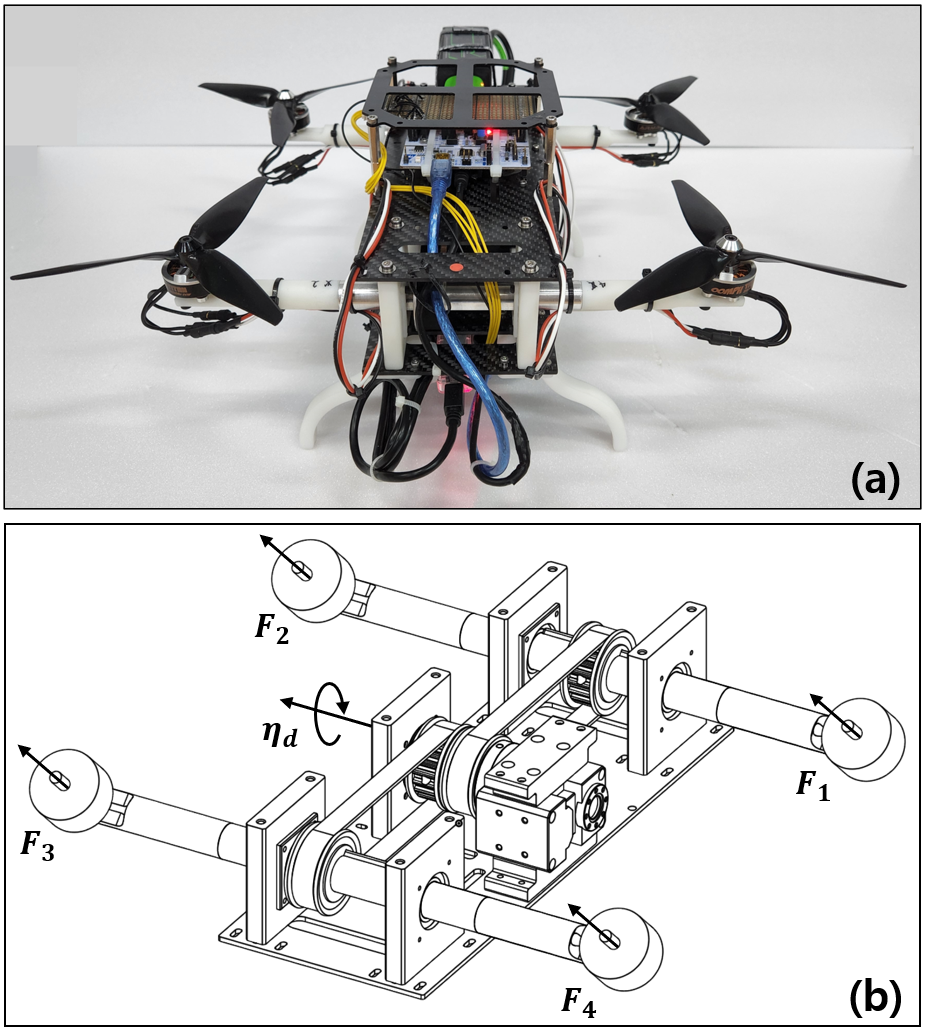}
    \caption{(a) Prototype of the proposed tiltrotor. (b) Illustration of control input $u = [\eta_d;F]$.}
    \label{fig:prototype and control input illustration}
    \vspace{-0.5cm}
\end{figure}

%We consider the following three requirements in hardware design: 1) only one additional actuator must be used, 2) all four rotors must be able to rotate at the same angle, and 3) the rotation angle must not be structurally obstructed. 
Based on these three considerations, we propose a tiltrotor with 5-CDoF whose prototype and CAD drawings can be found in Figs. \ref{fig:cad drawings} and \ref{fig:prototype and control input illustration}, respectively. We adopt a belt-pulley mechanism as can be found in Fig. \ref{fig:cad drawings}. Thanks to this mechanism, with only one additional actuator, we are able to rotate all four rotors by the same angle. Since no mechanical components obstructing the range of thrust vectoring exist, for example, a U-shaped component \cite{dynamixelPart} as installed in \cite{ding2021tilting,lee2021caros} or a linkage-related structure in \cite{ryll2022fast,lee2021fully}, the third requirement can be satisfied.

The added actuator is a servomotor which controls the angle of two rotor shafts which are illustrated in Fig. \ref{fig:cad drawings}. On each rotor shaft, two rotors are installed, by which the two rotors are mechanically constrained to rotate by the same angle of the rotor shaft. Next, the belt-pulley mechanism enforces the two rotor shafts to rotate by the same angle since a servo shaft is simultaneously constrained to both rotor shafts by belts and pulleys. Therefore, the two rotor shafts rotate by the same angle as the angle the servomotor rotates. For structural stability and smoothness in rotation, bearings are installed on the servo shaft and the two rotor shafts.

\section{Modeling}

Let $p \in \mathbb{R}^3$ be the center of mass of the tiltrotor in the world fixed frame and $\omega \in \mathbb{R}^3$ be the body angular velocity. We use the rotation matrix $R \in \text{SO}(3)$ and ZYX Euler angles $\phi = [\phi_1;\phi_2;\phi_3] \in \mathbb{R}^3$ to denote the orientation of the tiltrotor. Control input is regarded as $u = [\eta_d; F_1; F_2; F_3; F_4] \in \mathbb{R}^5$ where $\eta_d$ is the desired servomotor angle, and $F_i \in \mathbb{R}_+$ $i=1,\cdots,4$ is the rotor thrust. The control input is illustrated in Fig. \ref{fig:prototype and control input illustration} (b). For the ease of controller design, we consider a virtual control input of $\tilde{u}=[f_x;f_z;\tau] \in \mathbb{R}^5$ where $f_x,f_z \in \mathbb{R}$ are $x,z$-directional forces described in the body frame, and $\tau \in \mathbb{R}^3$ is a torque also in the body frame. Then, system dynamics of the proposed tiltrotor can be written as
\begin{subequations} \label{eq: system dynamics}
\begin{align}
    \begin{split} \label{eq: system dynamics - translation}
        m \ddot{p} &= R (f_x e_1 + f_z e_3) - m g e_3    
    \end{split} \\
    \begin{split} \label{eq: system dynamics - rotation}
        I_b \dot{\omega} &= -\omega \times I_b \omega + \tau    
    \end{split}
\end{align}
\end{subequations} 
where $m \in \mathbb{R}_+$ and $I_b \in \mathbb{R}^{3\times 3}$ are the mass and moment of inertia of the tiltrotor measured in the body frame. $g$ is a gravitational acceleration constant, and $a \times b$ denotes the cross product of $a,b \in \mathbb{R}^3$.

Before deriving a control law, it is essential to find a mapping between $u$ and $\tilde{u}$ so that we can calculate the actual control input $u$ whenever a control law for $\tilde{u}$ is established. Considering the kinematic configuration of the tiltrotor illustrated in Fig. \ref{fig:prototype and control input illustration}, the mapping between $u$ and $\tilde{u}$ is defined as $\tilde{u} = C(\eta) F$ where
\begin{equation*} 
\begin{gathered}
    C(\eta) = \\
    \resizebox{1.0\linewidth}{!}{$
        \left[\begin{matrix} s\eta & s\eta & s\eta & s\eta \\
                                 c\eta & c\eta & c\eta & c\eta \\
                                 -L_h c\eta-k_f s\eta & L_h c\eta +k_f s\eta & L_h c\eta-k_f s\eta & -L_h c\eta +k_f s\eta \\
                                 -L_v c\eta & -L_v c\eta & L_v c\eta & L_v c\eta \\
                                 L_h s\eta -k_f c\eta & -L_h s\eta +k_f c\eta & -L_h s\eta -k_f c\eta & L_h s\eta +k_f c\eta
        \end{matrix}\right]$}
\end{gathered}
\end{equation*}      
and $F = [F_1;F_2;F_3;F_4] \in \mathbb{R}^4$. $L_h, L_v \in \mathbb{R}_+$ are rotor-to-rotor distances divided by two in body $y$ and $x$ axes, respectively. $k_f \in \mathbb{R}_+$ denotes a thrust to torque ratio of a single rotor. Since servomotors show sufficiently fast tracking response in practice, i.e. $\eta \approx \eta_d$, we assume the following relationship for designing a controller: $\tilde{u} = C(\eta_d) F$.

\section{Controller Design} \label{sec:controller design}
\subsection{Control allocation}
A control allocation problem is to find a mapping from $\tilde{u}$ to $u$, which is the inverse of $u \mapsto \tilde{u}=C(\eta_d) F$. A closed-form solution for $u = [\eta_d;F]$ can be obtained as
\begin{equation} \label{eq: control allocation}
\begin{aligned}
\eta_d &= \text{atan2}(f_x,f_z) \\
F &= D(\eta_d) \left[\begin{matrix} \sqrt{f_x^2 + f_z^2} \\ \tau \end{matrix}\right]
\end{aligned}
\end{equation}
where 
\begin{equation*}
\begin{gathered}
    D(\eta_d) = \\
    \resizebox{1.0\linewidth}{!}{$\cfrac{1}{4}
        \left[\begin{matrix} 1 & -\frac{1}{k_f}s\eta_d-\frac{1}{L_h}c\eta_d & -\frac{1}{L_v c\eta_d} & -\frac{1}{k_f}c\eta_d+\frac{1}{L_h}s\eta_d \\
                            1 & \frac{1}{k_f}s\eta_d+\frac{1}{L_h}c\eta_d & -\frac{1}{L_v c\eta_d} & \frac{1}{k_f}c\eta_d-\frac{1}{L_h}s\eta_d \\
                            1 & -\frac{1}{k_f}s\eta_d+\frac{1}{L_h}c\eta_d & \frac{1}{L_v c\eta_d} & -\frac{1}{k_f}c\eta_d-\frac{1}{L_h}s\eta_d \\
                            1 & \frac{1}{k_f}s\eta_d-\frac{1}{L_h}c\eta_d & \frac{1}{L_v c\eta_d} & \frac{1}{k_f}c\eta_d+\frac{1}{L_h}s\eta_d 
        \end{matrix}\right]$}.
\end{gathered}
\end{equation*}
The matrix $D(\eta_d) \in \mathbb{R}^{4 \times 4}$ is well-defined if we restrict $\eta_d \in (-\pi/2,\pi/2)$ which can be assured if $f_z \neq 0$. Note that the fact of $D(\eta_d)$ not being able to be defined at $\eta_d = \pm \pi/2$, which occurs only when $f_z = 0$, is consistent with the hardware configuration since when $\eta_d (\approx \eta) = \pm \pi/2$, $\tau_2$ cannot be arbitrarily generated but zero.

\subsection{Controller design for $\tilde{u}$} \label{subsec: controller design for utilde}
Unlike a conventional multirotor, thanks to the tilting mechanism, the pitching motion of the tiltrotor can be independently controlled regardless of any translational motion thus enabling 5-CDoF motion. However, rolling motion cannot be independently controlled, and to consider this underactuatedness, we decompose the system dynamics into underactuated and fully actuated subsystems. Furthermore, since the control input $\tilde{u}$ is defined in the body frame, simply designing a controller in the world-fixed frame could result in singularity. Therefore, to resolve singularity in controller design, we design a controller using a new state variable $\tilde{p} = R_z(\phi_3)^\top p$ where $R_z(\theta) \in \text{SO}(3)$ denotes a rotation matrix with the angle $\theta$ along the body $z$-axis. 

\begin{remark}
If one naively chooses a configuration of the fully actuated subsystem as $[p_1;p_3;\phi] \in \mathbb{R}^5$ without resorting to a transformation such as the proposed one $\tilde{p} = R_z(\phi_3)^\top p$, singularity cannot be avoided. Since singularity in a control law typically leads to abrupt change of a control input, this property deteriorates control performance and thus should be avoided.
To investigate how singularity occurs with the naive choice of configurations, we first derive dynamics of $[p_1;p_3]$ which can be obtained from (\ref{eq: system dynamics - translation}) as
\begin{equation*}
    \left[\begin{matrix} \ddot{p}_1 \\ \ddot{p}_3 \end{matrix} \right] = \cfrac{1}{m} B(\phi) \left[\begin{matrix} f_x \\ f_z \end{matrix} \right] + \left[\begin{matrix} 0 \\ -g \end{matrix} \right]
\end{equation*}
where 
\begin{equation*}
    B(\phi) = \left[\begin{matrix} c\phi_2 c\phi_3 & s\phi_1 s\phi_3 + c\phi_1 s\phi_2 c\phi_3 \\ -s\phi_2 & c\phi_1 c\phi_2
    \end{matrix} \right].
\end{equation*}
Since $\text{det}(B) = c\phi_1 c\phi_3 + s\phi_1 s\phi_2 s\phi_3$, the input matrix $B(\phi)$ becomes singular whenever $\phi_3 = \pm \pi/2$ and $\phi_1$ or $\phi_2 =0$. Considering that matrix inverse of the input matrix $B$ is widely adopted in controller design, this singularity issue is detrimental to control performance, which could possibly lead to instability.

\end{remark}
\begin{comment}
Singularity occurs in the input matrix of the fully actuated subsystem. Rather than resorting to the proposed transformation $\tilde{p} = R_z(\phi_3)^\top p$, one may naively choose a configuration of the fully actuated subsystem as $[p_1;p_3;\phi] \in \mathbb{R}^5$. Then, the dynamics of $[p_1;p_3]$ can be obtained from (\ref{eq: system dynamics - translation}) as
\begin{equation*}
    \left[\begin{matrix} \ddot{p}_1 \\ \ddot{p}_3 \end{matrix} \right] = \cfrac{1}{m} B(\phi) \left[\begin{matrix} f_x \\ f_z \end{matrix} \right] + \left[\begin{matrix} 0 \\ -g \end{matrix} \right]
\end{equation*}
where 
\begin{equation*}
    B(\phi) = \left[\begin{matrix} c\phi_2 c\phi_3 & s\phi_1 s\phi_3 + c\phi_1 s\phi_2 c\phi_3 \\ -s\phi_2 & c\phi_1 c\phi_2
    \end{matrix} \right].
\end{equation*}
Since $\text{det}(B) = c\phi_1 c\phi_3 + s\phi_1 s\phi_2 s\phi_3$, we can conclude that the input matrix $B(\phi)$ becomes singular whenever $\phi_3 = \pm \pi/2$ and $\phi_1$ or $\phi_2 =0$. Considering that matrix inverse of the input matrix $B$ is widely adopted in controller design, this singularity issue is critical.
\end{comment}

\begin{figure}
    \centering
    \includegraphics[width=0.95\linewidth]{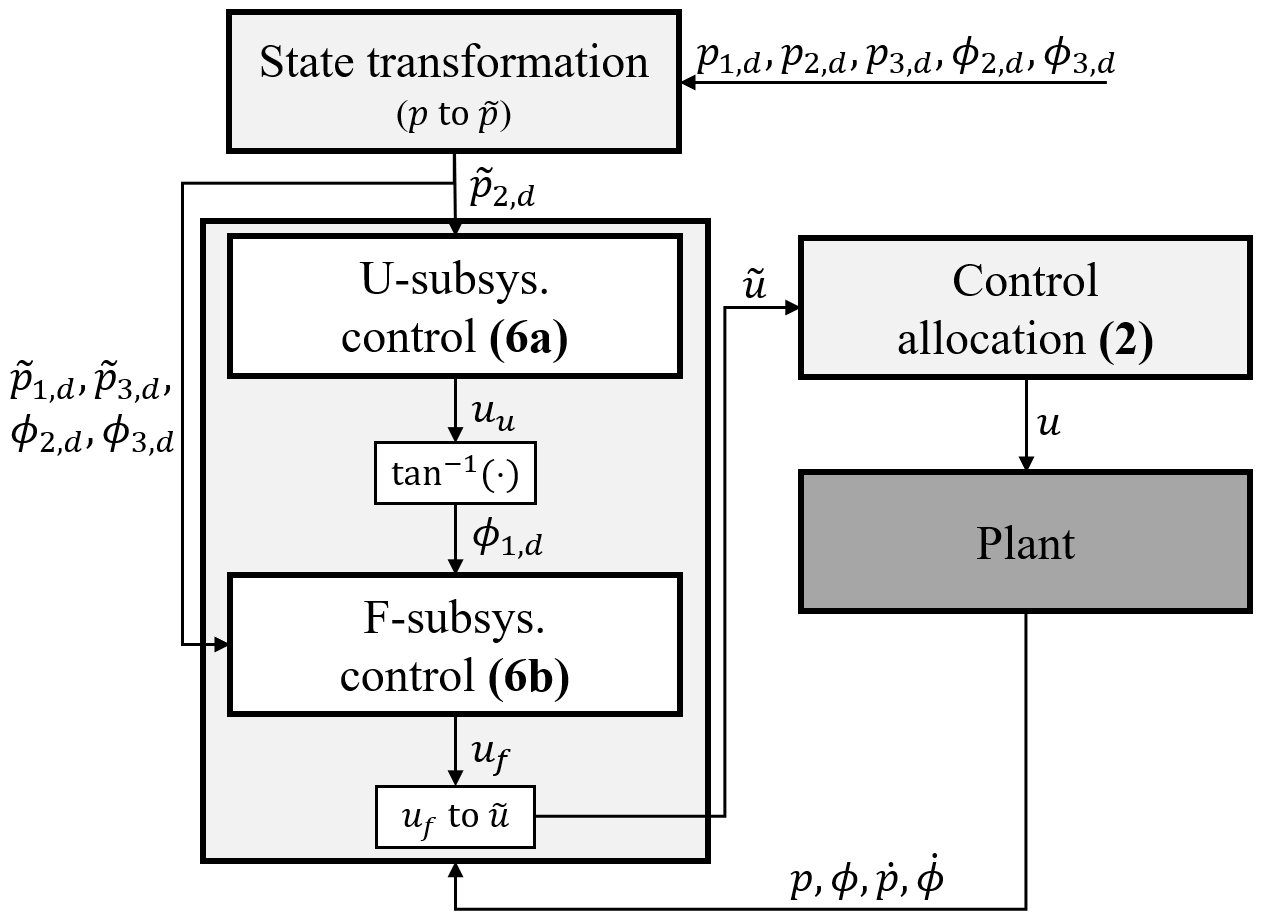}
    \caption{Flow chart of the proposed controller. As depicted as an input to the state transformation block, 5-CDoF (position $x,y,z$, pitch, and yaw) can be achieved by the proposed controller.}
    \label{fig:control flow chart}
    % \vspace{-0.5cm}
\end{figure}

The objective of a controller for $\tilde{u}$ is to track sufficiently smooth position $p_1,p_2,p_3$, pitch $\phi_2$, and yaw $\phi_3$ trajectories, or equivalently, transformed position coordinate $\tilde{p}_1,\tilde{p}_2,\tilde{p}_3$, pitch $\phi_2$, and yaw $\phi_3$ trajectories. However, due to underactuatedness in $\tilde{p}_2$, $\tilde{p}_2$ cannot be directly controlled with the control input $\tilde{u}$ but should be indirectly controlled with the roll angle $\phi_1$. Note that this control approach is similar to that of a conventional multirotor \cite{zhao2015nonlinear} whose $x,y$ position is controlled with roll and pitch angles and not directly with a control input. To this end, we propose a cascade control law where we first control $\tilde{p}_2$, which we call an underactuated subsystem, to obtain a desired roll angle $\phi_{1,d}$, and the rest configuration $[\tilde{p}_1;\tilde{p}_3;\phi_1;\phi_2;\phi_3] \in \mathbb{R}^5$ entitled fully actuated subsystem is controlled afterward. This cascade control framework is illustrated in Fig. \ref{fig:control flow chart}.

Let $Q \in \mathbb{R}^{3 \times 3}$ denote a Jacobian matrix satisfying $\omega = Q \dot{\phi}$, then rotational dynamics with respect to $\phi$ can be derived from (\ref{eq: system dynamics - rotation}) as
\begin{equation} \label{eq: rotational dynamics - RPY}
    \ddot{\phi} = h_\phi + g_\phi \tau   
\end{equation}
where $h_\phi = (I_b Q)^{-1}\{-I_b \dot{Q} \dot{\phi} - \omega \times (I_b \omega)\}$ and $g_\phi = (I_b Q)^{-1}$.
Dynamics with respect to $\tilde{p}$ can be computed from (\ref{eq: system dynamics - translation}) as
\begin{equation} \label{eq: translational dynamics - transformation}
    \ddot{\tilde{p}} = \frac{1}{m}R_y(\phi_2) R_x(\phi_1) (f_x e_1 + f_z e_3) - ge_3 + h_p + g_p \tau
\end{equation}
where $R_x(\theta)$, $R_y(\theta)$ $\in \text{SO}(3)$ denote $x,y$-directional rotation matrices with the angle $\theta$. $h_p \in \mathbb{R}^3$, $g_p \in \mathbb{R}^{3\times3}$ are terms resulting from state transformation and are defined as follows
\begin{equation*}
\begin{aligned}
    h_p &= -\hat{e}_3 R_z(\phi_3)^\top (2 \dot{\phi}_3 \dot{p} + e_3^\top h_\phi p) - \hat{e}_3^2 R_z(\phi_3)^\top \dot{\phi}_3^2 p \\
    g_p &= -\hat{e}_3 R_z(\phi_3)^\top p e_3^\top g_\phi.
\end{aligned}
\end{equation*}
$\hat{(\cdot)}$ is a hat operator mapping a vector in $\mathbb{R}^3$ to a skew-symmetic matrix in $\mathbb{R}^{3 \times 3}$. Then, to decompose underactuated and fully actuated dynamics, we define their configurations as $q_u = \tilde{p}_2 \in \mathbb{R}$ and $q_f = [\tilde{p}_1;\tilde{p}_3;\phi] \in \mathbb{R}^5$, respectively. Defining $\bar{f}_z 
\coloneqq c\phi_1 f_z$, both dynamics can be arranged as follows:
\begin{subequations} \label{eq: underactuated and fully actuated dynamics}
\begin{align}
    \begin{split}
        \ddot{q}_u &= h_u + g_u u_u - \tfrac{1}{m}(t\phi_1 - t\phi_{1,d}) \bar{f}_z
    \end{split}\\
    \begin{split}
        \ddot{q}_f &= h_f + g_f u_f
    \end{split}
\end{align}
\end{subequations}
where $h_u = e_2^\top (h_p+g_p \tau) \in \mathbb{R}$, $g_u = -\frac{1}{m}\bar{f}_z \in \mathbb{R}$, $h_f = [e_1^\top h_p; e_3^\top h_p; h_\phi] \in \mathbb{R}^5$, and 
\begin{equation*}
    g_f = \left[\begin{matrix} \frac{1}{m} \left[\begin{matrix} c\phi_2 & s\phi_2 \\ -s\phi_2 & c\phi_2 \end{matrix} \right] & \left[\begin{matrix} e_1^\top \\ e_3^\top \end{matrix} \right] g_p \\
    0_{3\times 2} & g_\phi
    \end{matrix} \right] \in \mathbb{R}^{5 \times 5}.
\end{equation*}
$u_u \coloneqq t\phi_{1,d}$ and $u_f \coloneqq [f_x;\bar{f}_z;\tau]$ are virtual control inputs to each subsystem. For $\phi_1,\phi_2 \in (-\pi/2,\pi/2)$, $g_f$ is always invertible, and $f_z$ exists for any given $\bar{f}_z$. Then, with error states $e_u \coloneqq q_u - q_{u,d}$ and $e_f \coloneqq q_f - q_{f,d}$, we propose a control law as 
\begin{subequations} \label{eq: control law}
\begin{align}
    \begin{split}
        u_u = g_u^{-1} \left( \ddot{q}_{u,d} -h_u - K_{up} e_u - K_{ud} \dot{e}_u - K_{ui} \int e_u dt \right)
    \end{split} \\
    \begin{split}
        u_f = g_f^{-1} \left( \ddot{q}_{f,d} -h_f - K_{fp} e_f - K_{fd} \dot{e}_f - K_{fi} \int e_f dt \right).
    \end{split}
\end{align}
\end{subequations}

\begin{proposition} \label{prop1}
For positive-definite, diagonal matrices $K_{up}$, $K_{ud}$, $K_{ui}$, $K_{fp}$, $K_{fd}$, $K_{fi}$ satisfying $K_{up}K_{ud} > K_{ui}$ and $K_{fp}K_{fd} > K_{fi}$, the closed-loop system consisting of the system dynamics (\ref{eq: underactuated and fully actuated dynamics}) and control input (\ref{eq: control law}) is asymptotically stable.
\end{proposition}
\begin{proof}
The closed-loop system can be written as
\begin{subequations} 
\begin{align}
    \begin{split} \label{eq: closed-loop system - underactuated}
        \dddot{e}_u + K_{ud} \ddot{e}_u + K_{up} \dot{e}_u &+ K_{ui} e_u = \\
        &-\frac{1}{m} \frac{d}{dt}\left\{(t\phi_1 - t\phi_{1,d}) \bar{f}_z \right\}
    \end{split} \\
    \begin{split} \label{eq: closed-loop system - fully actuted}
        \dddot{e}_f + K_{fd} \ddot{e}_f + K_{fp} \dot{e}_f &+ K_{fi} e_f = 0.
    \end{split}
\end{align}
\end{subequations}
By examining Routh-Hurwitz criterion with the conditions on controller gains, (\ref{eq: closed-loop system - fully actuted}) is exponentially stable to $[e_f;\dot{e}_f] = 0$ and (\ref{eq: closed-loop system - underactuated}) is also exponentially stable to $[e_u;\dot{e}_u] = 0$ if $[e_f;\dot{e}_f] = 0$. Therefore, by stability theorem for a cascade system \cite{seibert1990global}, which was also adopted in \cite{lee2022rise,kim2018stabilizing}, the entire system is asymptotically stable.
\end{proof}

Thanks to the above \textit{Proposition} \ref{prop1}, asymptotic stability with respect to the all error variables $e_u,e_f$ is guaranteed with full consideration of underactuatedness. Therefore, we can confirm that the original objective of the controller, which is to track sufficiently smooth position $p_1,p_2,p_3$, pitch $\phi_2$, and yaw $\phi_3$ trajectories, is accomplished.
Overall flow chart of the proposed controller can be found in Fig. \ref{fig:control flow chart}. $u_f$ to $\tilde{u}$ conversion can be easily computed by applying the definition of $\bar{f}_z = c\phi_1 f_z$. 

\section{Results}
\subsection{Experimental setup}
As in Fig. \ref{fig:prototype and control input illustration}, mechanical components of the proposed tiltrotor are two rotor shafts and one servo shaft all made up of aluminum, four 3D printed rotor stands connecting the shafts and the rotors, two carbon plates, five 3D printed bearing holders, five aluminum bearings, four timing pulleys, and two timing belts. The distance between one rotor shaft to the other is $0.196 \text{ m}$, and the distance between the two rotors attached to the same rotor shaft is $0.257 \text{ m}$. Overall mass is about $2.1 \text{ kg}$ including one 4S 2200 mAh LiPo battery. For actuators and related components, we adopt four Armattan 2306/2450KV rotors, four APC 6-inch propellers, one HOBBYWING XRotor Micro 60A 4in1 ESC, and one ROBOTIS Dynamixel XM430-W350. For localization, OptiTrack motion capture system and LORD 3DM-GX3-25 IMU are used whose signals are processed with error state Kalman filter \cite{sola2017quaternion}. To generate PWM signals to ESC, STM NUCLEO-F446RE board is used. We harness Intel NUC running Robot Operating System (ROS) in Ubuntu 20.04 as an onboard computer which executes all algorithms including localization and control. 

To first validate the proposed hardware platform and controller, the following two scenarios are conducted: 1) $x,y$ position tracking control while maintaining zero pitch angle and 2) non-zero pitch tracking control with no translation. We select these two scenarios because they could illustrate maneuvers that cannot be achieved with a conventional quadrotor. Lastly, to demonstrate perching-enabled APhI with the tiltrotor, we conduct a scenario of perching and cart pushing. In Figs. \ref{fig:exp_result_xyTracking_plot}, \ref{fig:exp_result_pitchTilt_plot}, \ref{fig:exp_result_cartPushing_plot}, we plot time history of actual state/input signals with solid lines while that of desired state signals with dashed lines.

\subsection{Scenario 1 -- position tracking control without pitching motion}

\begin{figure}
    \centering
    \includegraphics[width=1.0\linewidth]{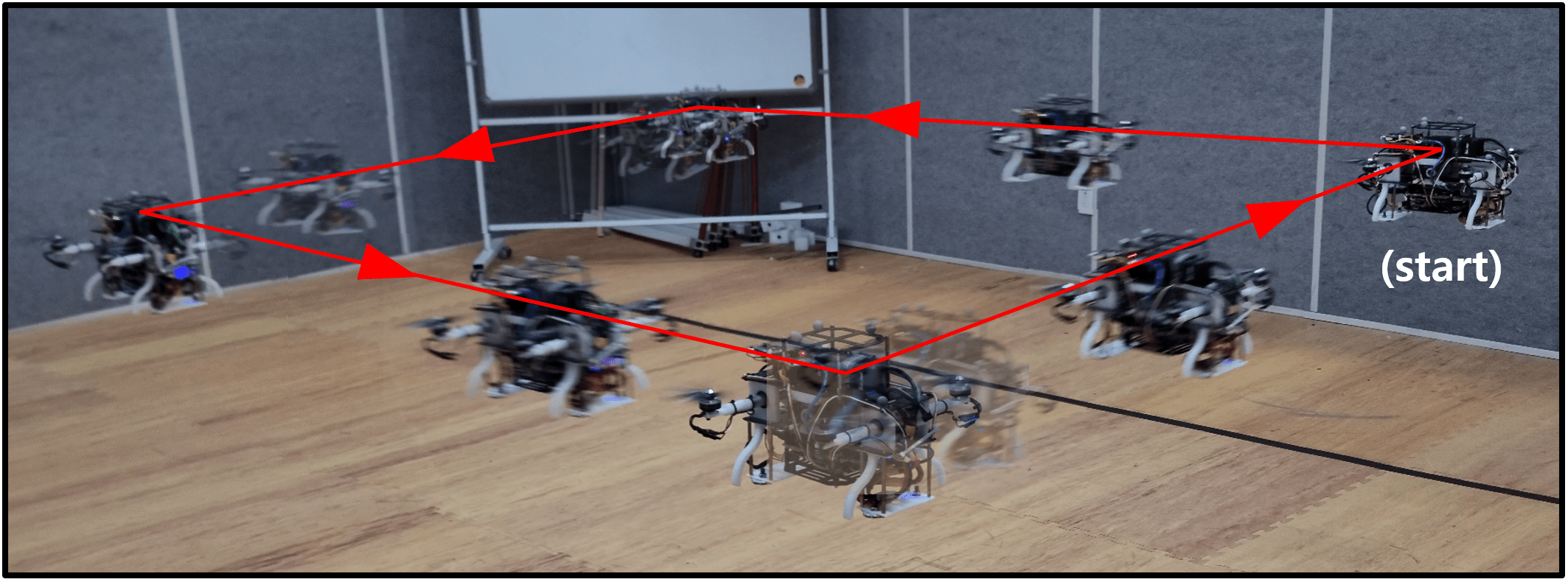}
    \caption{A composite image of a tiltrotor tracking a square trajectory in $XY$-plane.}
    \label{fig:exp_result_xyTracking}
\end{figure}
\begin{figure}
    \centering
    \includegraphics[width=1.0\linewidth]{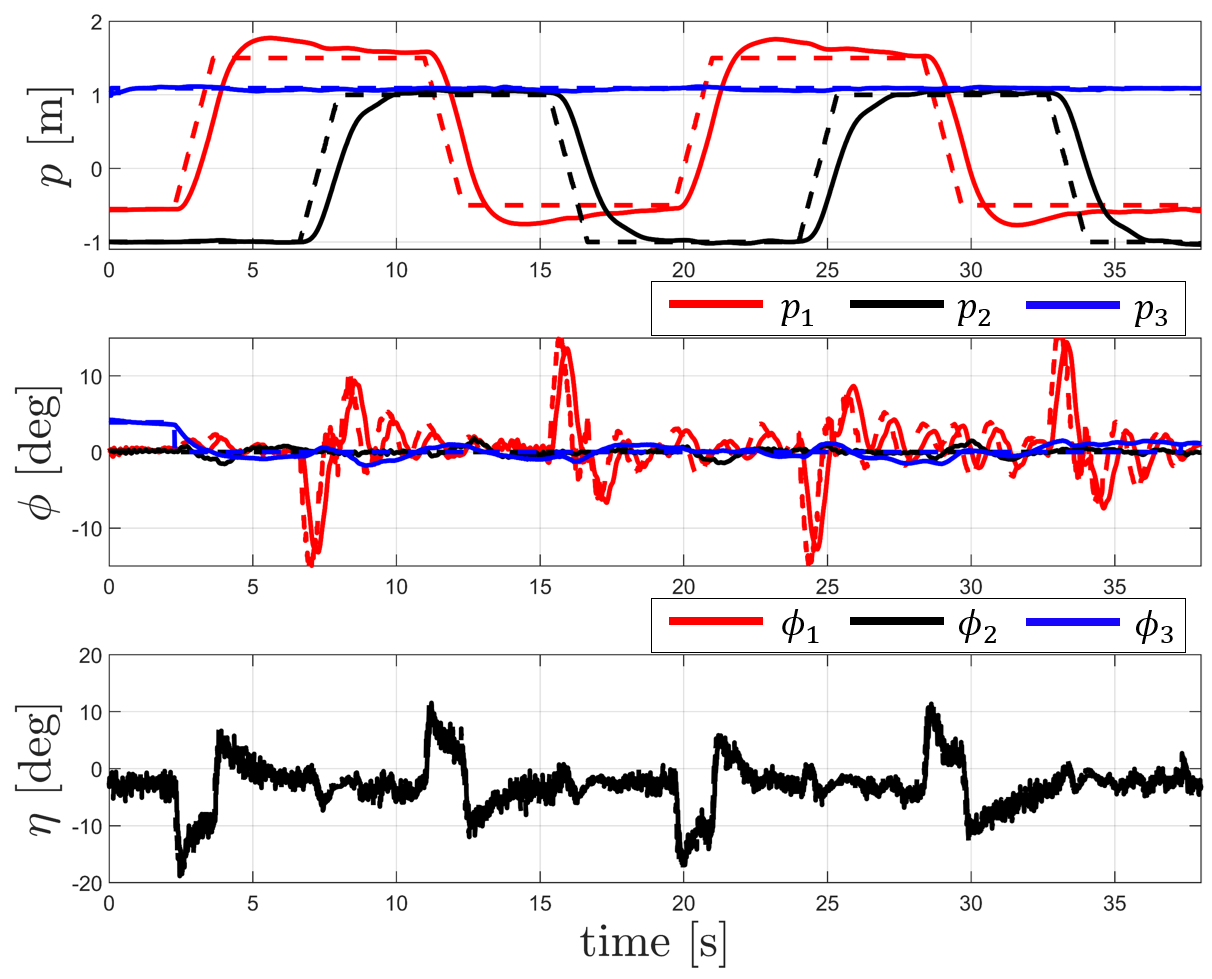}
    \caption{Time history of the states during the experiment of tracking square trajectory in $XY$-plane.}
    \label{fig:exp_result_xyTracking_plot}
\end{figure}

%The first two scenarios are to show that the tiltrotor can actually attain 5-CDoF. 
In the first scenario, we show that $x$-directional translation can be independently controlled without pitching motion. For comparison, we also let the tiltrotor translate in $y$-direction whose motion can only be achieved with rolling motion due to underactuatedness in $y$-direction. As can be found in Figs. \ref{fig:exp_result_xyTracking}, \ref{fig:exp_result_xyTracking_plot}, we define a square reference trajectory in $XY$-plane (red and black dashed lines at the top of Fig. \ref{fig:exp_result_xyTracking_plot}) and let pitch and yaw desired values to be uniformly zero (blue and black dashed lines in the middle of Fig. \ref{fig:exp_result_xyTracking_plot}). Compared to the actual roll angle and its desired value which can be found as red solid and dashed lines in the middle figure, almost no pitching motion occurs, and thus we can confirm independent controllability in the $x$-axis. In the bottom figure of Fig. \ref{fig:exp_result_xyTracking_plot}, we can find that the added servomotor rotates to keep pitch angle zero while tracking the time-varying $x$-directional reference.

\subsection{Scenario 2 -- pitch tracking control without translation}
\begin{figure}
    \centering
    \includegraphics[width=0.90\linewidth]{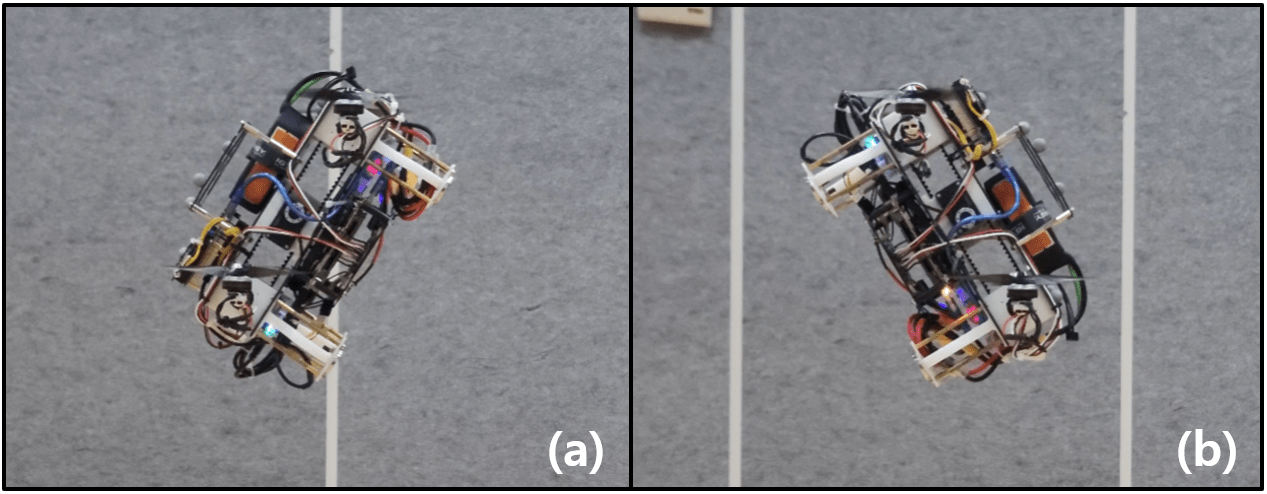}
    \caption{Captured images of a tiltrotor tracking time-varying, non-zero pitch angle. ((a): 60 deg. (b): -60 deg.)}
    \label{fig:exp_result_pitchTilt}
\end{figure}

\begin{figure}
    \centering
    \includegraphics[width=1.0\linewidth]{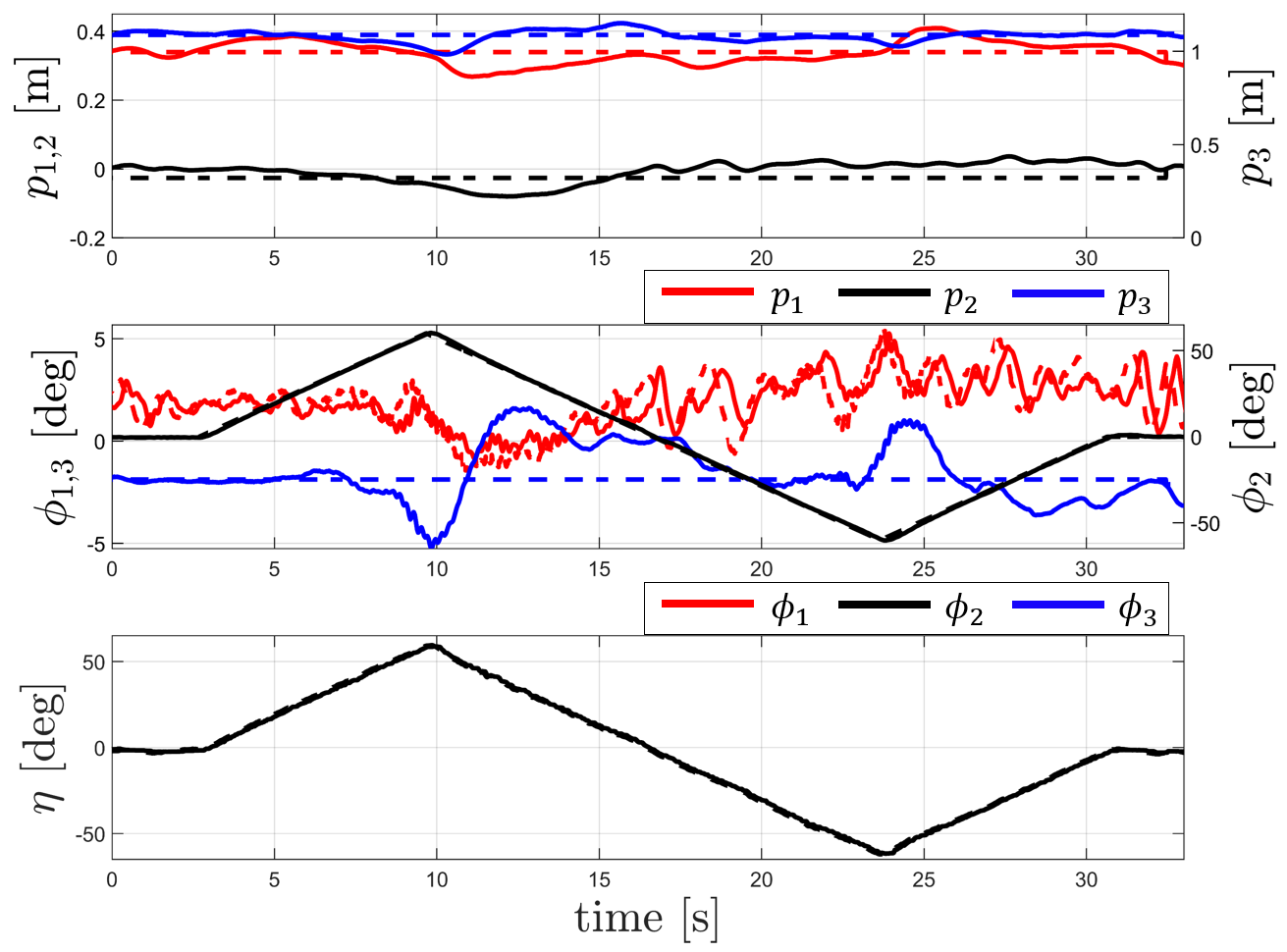}
    \caption{Time history of the states during the experiment of tracking time-varying, non-zero pitch angle ($\phi_{2,d} \in [-60,60] \text{ [deg]}$).}
    \label{fig:exp_result_pitchTilt_plot}
\end{figure}

The second scenario is to show independent controllability in the pitching motion. To show this, we let the desired trajectory of $x$ be constant (the red dashed line at the top of Fig. \ref{fig:exp_result_pitchTilt_plot}) and let the desired pitch angle be time-varying (the black dashed line in the middle of Fig. \ref{fig:exp_result_pitchTilt_plot}). As can be found in Figs. \ref{fig:exp_result_pitchTilt} and \ref{fig:exp_result_pitchTilt_plot}, the tiltrotor could rotate in pitch direction while regulating its position. About $\pm 60 \text{ deg}$ is achieved during the experiment, and even at those states, the tiltrotor could stably hover.


\subsection{Scenario 3 -- perching-enabled aerial physical interaction}
\begin{figure}
    \centering
    \includegraphics[width=0.75\linewidth]{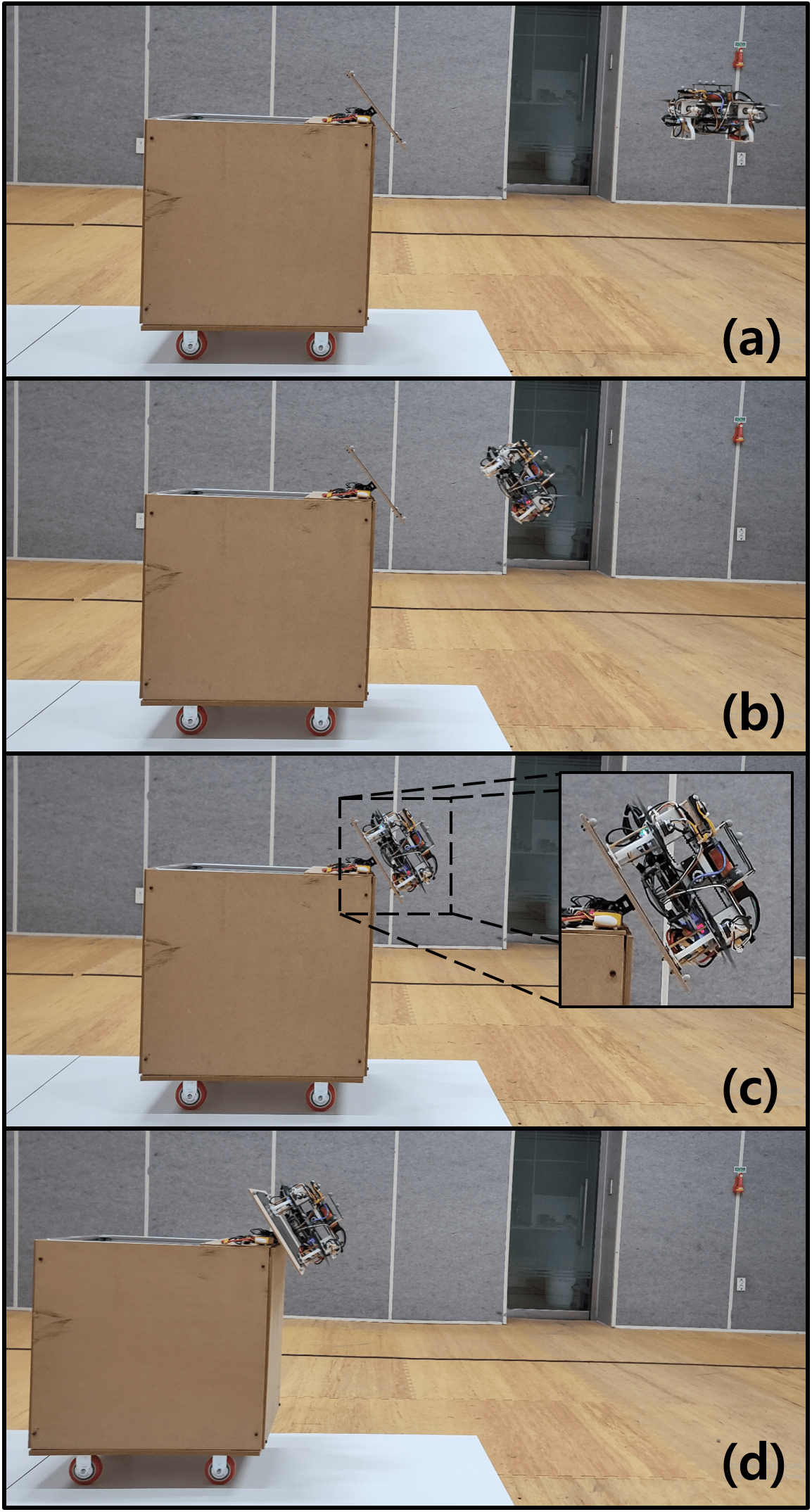}
    \caption{Captured images of a perching-enabled cart pushing. The time sequence is indicated by the alphabetic order, (a) to (d).}
    \label{fig:exp_result_cartPushing}
\end{figure}
\begin{figure}
    \centering
    \includegraphics[width=1.0\linewidth]{figures/cartPushing_plot_v3.PNG}
    \caption{Time history of the states and normalized PWM signal to each rotor during the experiment of perching-enabled cart pushing. The gray shaded region indicates the time after perching.}
    \label{fig:exp_result_cartPushing_plot}
\end{figure}

Before performing a perching-enabled APhI, we assume that the pose of the perching site is known and that perching is passively conducted. Although we utilize the Optitrack motion capture system to satisfy the first assumption, this can also be done in a fully onboard manner by adopting an onboard sensor (e.g. camera). To ensure the second assumption, we use magnets. However, this assumption can also be alleviated by adopting other connecting mechanisms (e.g. grasping or adhesion) \cite{wopereis2016mechanism,popek2018autonomous,tsukagoshi2015aerial} but we believe this is beyond the scope of the current study.

We apply a simple rule-based planning algorithm to make the tiltrotor navigate from the initial pose to the perching site. A linearly interpolated reference trajectory from the initial pose to the perching site is computed in the planning algorithm where the interpolation is conducted sequentially only in one axis at a time based on the sequence of $z \rightarrow y \rightarrow yaw \rightarrow pitch \rightarrow x$. As can be found in Figs. \ref{fig:exp_result_cartPushing}, \ref{fig:exp_result_cartPushing_plot}, perching can be successfully accomplished even with such simple planning method thanks to the 5-CDoF property. After perching, which is indicated by the gray shaded region in Fig. \ref{fig:exp_result_cartPushing_plot}, we manually switch the controller to perching-enabled APhI mode where the thrust direction rotates to be perpendicular to the perching site surface. Then, the PWM values of all four rotors are increased simultaneously to make the cart move. This PWM increase can be found at the bottom of Fig. \ref{fig:exp_result_cartPushing_plot}, and the resulting cart motion can be found in the red solid line at the top figure representing the $x$-directional position of the tiltrotor.

\section{Conclusion and future work}

In this study, we proposed a hardware design of a minimally actuated 5-CDoF tiltrotor and an asymptotically stabilizing controller for controlling the full 5-CDoF. Features of the proposed tiltrotor are that 1) it is minimally actuated, 2) it has a high interaction force margin during aerial physical interaction (APhI), and 3) no mechanical obstruction exists in thrust direction rotation. These properties allow the platform to be specialized for perching-enabled APhI because it is capable of parallel hovering to any inclined surface and thrust direction tilting without any hurdle. Then, to handle singularity and underactuatedness in the controller design, we designed a controller based on the transformed coordinate and the decomposed subsystems. The asymptotic stability of the entire system was then proved without any assumption on underactuatedness. To validate the proposed platform and the controller, we conducted two experiments to show the performance of controlling the extra CDoF compared to a conventional quadrotor. Furthermore, to demonstrate applicability of the proposed platform to perching-enabled APhI, a perching and cart pushing experiment was successfully conducted. 
As a future work, we expect to modify the hardware design to enable fully vertical hovering and perching. This can be accomplished by providing an offset between the two tilting axes in the body $z$ direction, which does not violate any hardware requirement assigned in this paper. Along with this hardware modification, we will expand this study to full pose control of a movable object using multiple perching-enabled tiltrotors as normal force generating modules.

%automate the whole process of perching-enabled APhI and design a robust controller with strict stability analysis for the perturbed system.

\addtolength{\textheight}{-12cm}   % This command serves to balance the column lengths
\normalem % to prevent the effect from \usepackage{ulem}
% \clearpage
% \bibliographystyle{./bibtex/IEEEtran}
% \bibliography{./bibtex/IEEEabrv, ./bibtex/mybibfile}

\end{document}